\documentclass[12pt,A4]{article}
\usepackage[margin=1in,footskip=0.25in]{geometry}
\usepackage{multicol}
\usepackage{graphicx, color}  
\usepackage{amsmath,amssymb, amsfonts, MnSymbol, mathrsfs}
\usepackage{natbib}
\usepackage{hyperref}
\usepackage{rotating} 
\usepackage{tikz}

  \def\ni{\noindent}

\begin{document}

\title{Human-Calibrated Automated Testing and Validation of Generative Language Models: An Overview}
\author{Agus Sudjianto$^{1,2}$, Aijun Zhang$^{3,*}$, Srinivas Neppalli$^1$,\\ Tarun Joshi$^{3,*}$, Michal Malohlava$^1$} 
\renewcommand{\thefootnote}{}
\footnotetext{$^1$H2O.ai, $^2$Center for Trustworthy AI Through Model Risk Management, UNC Charlotte, $^3$Wells Fargo.}
\footnotetext{$^*$The views expressed in this paper are those of the authors and do not necessarily reflect those of Wells Fargo.}
\date{}
\maketitle

\begin{abstract}
This paper introduces a comprehensive framework for the evaluation and validation of generative language models (GLMs), with a focus on Retrieval-Augmented Generation (RAG) systems deployed in high-stakes domains such as banking. GLM evaluation is challenging due to open-ended outputs and subjective quality assessments. Leveraging the structured nature of RAG systems, where generated responses are grounded in a predefined document collection, we propose the Human-Calibrated Automated Testing (HCAT) framework. HCAT integrates a) automated test generation using stratified sampling, b) embedding-based metrics for explainable assessment of functionality, risk and safety attributes, and c) a two-stage calibration approach that aligns machine-generated evaluations with human judgments through probability calibration and conformal prediction.   

In addition, the framework includes robustness testing to evaluate model performance against adversarial, out-of-distribution, and varied input conditions, as well as targeted weakness identification using marginal and bivariate analysis to pinpoint specific areas for improvement. This human-calibrated, multi-layered evaluation framework offers a scalable, transparent, and interpretable approach to GLM assessment, providing a practical and reliable solution for deploying GLMs in applications where accuracy, transparency, and regulatory compliance are paramount.

\vskip 6.5pt \noindent {\bf Keywords}: Generative Language Models, Retrieval-Augmented Generation, Model Validation, Human-Calibrated Testing, Automated Test Generation, Embedding-Based Metrics, Conformal Prediction, Robustness Testing, Weakness Identification.
\end{abstract}

\section{Introduction}
Generative language models (GLMs) have revolutionized natural language processing (NLP), powering applications such as conversational agents, content generation, and language translation. The latest large language models (LLMs) \citep{brown2020language} can generate coherent and contextually relevant text that often closely resembles human writing. Retrieval-augmented generation (RAG) systems advance this capability by embedding retrieval mechanisms within generative models, enabling access to a structured knowledge base of documents to guide responses \citep{krishna2023rag, lewis2020retrieval}. In RAG systems, this defined collection of documents serves as a grounding reference, allowing the generation process to produce outputs anchored in available information. This approach helps ensure more predictable and manageable system behavior by constraining outputs within a predefined document scope.

Evaluating GLMs is challenging due to the vast range of potential inputs and outputs, making exhaustive manual assessment impractical \citep{Srivastava2023, Liang2023HELM}. However, the bounded nature of RAG systems offers opportunities for more focused and feasible testing. With a defined document scope, RAG systems enable systematic exploration of inputs and provide a clearer framework for output evaluation. Additionally, the reliance on external documents allows the generation of relevant queries and anticipated responses, facilitating a more automated approach to testing and validation.

\subsection{Challenges in Evaluating Generative Models in Banking}
Model testing and validation is a rigorous process aimed at identifying and quantifying weaknesses in models to enable targeted improvements or to apply risk mitigation. This process ensures that models are reliable, accurate, and effective for their intended uses. Model validation helps prevent potential failures and maintains confidence in the model performance for critical business process. Banking industry in the US probably has the most matured model validation practice compared to other industries where every model prior going to production must be evaluated in terms of conceptual soundness and outcome analysis \citep{sudjianto2024MVP}.

Model testing and validation are essential to identify and quantify model weaknesses, enabling targeted improvements or risk mitigation measures. This process ensures that models remain reliable, accurate, and effective for their intended applications. In the banking sector, where model performance is critical to business operations, model validation practices are highly advanced. The U.S. banking industry, in particular, has established some of the most rigorous model validation standards, requiring every model to undergo comprehensive evaluation—including assessments of conceptual soundness and outcome analysis—before deployment \citep{sudjianto2024MVP}. However, generative language models  present unique challenges for validation compared to traditional predictive models, whose outputs are typically constrained to specific labels or numerical values. GLMs, in contrast, produce open-ended text, making it difficult to define a singular "correct" output and to consistently assess quality.

RAG systems combine retrieval and generation capabilities to produce nuanced, contextually rich responses. This dual functionality requires evaluation across several dimensions, including the relevance of retrieved information, the groundedness of generated content in those sources, the completeness of responses to user queries, and the overall relevance of the provided answers. As a result, evaluating RAG systems is inherently more complex than assessing traditional predictive models with well-defined outputs.

An emerging approach to evaluating GLMs involves using LLMs as judges to assess responses from other models based on metrics such as truthfulness, relevance, and consistency. For example, the TruthfulQA benchmark developed by \cite{Lin2022TruthQA} employs a fine-tuned LLM to evaluate responses for factual accuracy. While efficient and scalable, this "LLM-as-judge" method introduces limitations, including circularity and shared biases—where the evaluating LLM may have similar misconceptions or predispositions as the models it assesses. Additionally, in regulated industries like banking, relying on LLMs for evaluation may face resistance due to the opacity and lack of transparency in how outcomes are determined, making it challenging to meet explainability requirements. Further, using LLMs as evaluators may raise concerns about conceptual soundness, as evaluation by a complex, often unexplainable model might not align with rigorous validation standards expected in banking. The HELM (Holistic Evaluation of Language Models) framework proposed by \cite{Liang2023HELM} highlights the need for multidimensional assessments covering robustness, fairness, and toxicity, which are difficult to achieve reliably through opaque models. These issues underscore the broader challenge of adopting LLM-based evaluation methods in regulated sectors where transparency, conceptual soundness, and accountability are paramount for regulatory acceptance.

In summary, the evaluation of RAG systems and generative language models (GLMs) in banking presents distinct challenges. First, comprehensive testing is essential to ensure model reliability, yet the open-ended nature of GLMs makes defining exhaustive test cases difficult. Second, scaling this comprehensive testing is a formidable task, as test cases must be generated across a wide array of scenarios to capture the complexity of potential inputs and outputs. Third, establishing reliable evaluation approaches and metrics that can consistently capture dimensions like truthfulness, relevance, and groundedness is challenging, especially given the subjective nature of language quality. Finally, applying rigorous testing and validation procedures akin to those used in traditional predictive models is difficult for GLMs, as these models require assessment of open-text outputs rather than discrete, predictable values. Together, these challenges underscore the need for innovative testing methodologies and robust, scalable validation frameworks that can address the unique complexities of generative models in high-stakes industries like banking.

\subsection{A Structured Approach}
To address the challenges, we propose a structured approach covering essential steps for generative language model validation in banking:
\begin{enumerate}
    \item {\sf Define Model Purpose and Scope}:
    Begin by clearly stating the model’s intended use, whether for customer support, document summarization, or other applications. Establish boundaries around the expected tasks and the limitations of the model to guide validation criteria.
    \item {\sf Identify Potential Risks and Failures}:
    Outline the key risks, such as the model generating incorrect, biased, or misleading responses, which could have regulatory or reputational impacts. Focus on identifying failure modes specific to the financial and non-financial contexts, including the generation of non-compliant or sensitive information.
    \item {\sf Develop Diverse Test Cases and Stress Tests}:
    Create test cases that cover a broad spectrum of query types, ensuring that the model can handle varied topics, question formats, and levels of complexity. Conduct stress tests, such as ambiguous, adversarial, and out-of-distribution inputs, to reveal model weaknesses and test its robustness under challenging scenarios.
    \item {\sf Use Transparent and Explainable Metrics}:
    Prioritize metrics that offer transparency, such as those based on semantic similarity and natural language inference, over black-box methods. Embedding-based metrics like BERTScore or entailment probabilities can provide more interpretable insights into whether the model responses are relevant, grounded, and complete.
    \item {\sf Automate Testing for Comprehensive Coverage}:
    Automate testing when possible to ensure the validation process is scalable and can cover a wide range of queries and scenarios without requiring extensive manual effort.
    \item {\sf Calibrate with Human Evaluations}:
    Periodically sample model outputs for human review, comparing automated metrics with human judgments to ensure the automated processes align well with human expectations. Use this calibration to adjust metric thresholds or test parameters as needed.
    \item {\sf Identify Weaknesses for Model Improvement and Risk Mitigation}:
    Based on the validation results, highlight areas where the model struggles, such as specific query types or scenarios. Use this analysis to guide ongoing model improvement, risk mitigation or guardrails, and the design of monitoring systems to maintain performance post-deployment.
\end{enumerate}

This structured approach provides a comprehensive framework for validating GLMs within banking, integrating rigorous steps to ensure reliability, accuracy, and compliance with regulatory standards. 
Transparent, explainable metrics are prioritized to offer interpretable insights into GLM outputs, and automation is employed to enable thorough coverage across various scenarios. Calibration with human evaluations further aligns the validation process with real-world expectations, and continuous monitoring of GLM weaknesses ensures targeted improvements and mitigates risks over time.

\subsection{Human-Calibrated Automated Testing (HCAT) Framework}
Building on the structured approach outlined above, we now introduce the human-calibrated automated testing (HCAT) framework, a technical and systematic solution tailored for the rigorous demands of GLM testing and validation. 
This framework combines automated test generation, explainable evaluation metrics, and human-calibrated benchmarks to tackle the complexities of assessing GLMs, particularly in the context of RAG systems.

The HCAT framework is designed to ensure that the validation process is both scalable and interpretable, meeting high standards of transparency, accuracy, and compliance. The following components define the technical structure of the HCAT framework:
\begin{enumerate}
    \item {\sf Automatic Test Generation}: Using topic modeling and stratified sampling, HCAT produces a diverse set of queries covering the full scope of the document collection, enabling comprehensive model evaluation across varied input scenarios.
    \item {\sf Explainable Evaluation Metrics}: HCAT employs embedding-based metrics to provide a holistic assessment of model performance, spanning two critical dimensions:
    \begin{itemize}
    \item {\sf Functionality Metrics}: Embedding-based metrics assess core RAG capabilities, including relevance, groundedness, completeness, and answer relevancy, offering transparent and interpretable insights into semantic alignment between queries, contexts and answers.
    \item {\sf Risk and Safety Metrics}: Specialized embedding-based metrics assess risk and safety, such as toxicity, bias, and privacy protection, crucial for ensuring compliance and reliability in sensitive applications.
    \end{itemize}
    \item {\sf Calibration with Human Judgments}: To ensure that the automated metrics align with human perceptions, we calibrate them using samples of human labeling. This process involves:
    \begin{itemize}
    \item {\sf Sampling Human Evaluations}: Gathering human judgments on subsets of the generated outputs.
    \item {\sf Regression Techniques}: Applying probability calibration models to align machine evaluation scores with human judgments. 
    \item {\sf Conformal Prediction}: Quantifying uncertainty in machine evaluations by providing prediction sets with confidence level, enabling a more nuanced understanding of evaluation reliability.
    \end{itemize}
\end{enumerate}

In the following sections, we provide a detailed breakdown of each HCAT component. Section~\ref{sec:autotest} describes the automatic test generation process, including the use of topic modeling and stratified sampling to create a comprehensive set of test queries. Section~\ref{sec:funcmetrics} delves into functionality evaluation metrics, covering relevance, groundedness, completeness, and answer relevancy, and explains the use of embedding-based metrics to assess each dimension. Section~\ref{sec:riskmetrics} focuses on risk and safety evaluation, detailing metrics for toxicity, bias, and privacy protection to ensure compliance and reliability. Section~\ref{sec:calib} addresses the calibration process with human judgments, explaining how human evaluations refine automated metrics for real-world alignment. Section~\ref{sec:robustweak} presents robustness testing and weakness identification techniques to pinpoint areas for improvement, followed by Section~\ref{sec:con} with a discussion of implications and conclusions.

\section{Automatic Test Generation}\label{sec:autotest}
To evaluate GLMs comprehensively, particularly RAG systems, it is crucial to have a diverse and representative set of queries that spans the entire scope of the document collection.  To achieve this, we propose an automatic query generation method through stratified sampling. The topic modeling technique by \cite{Gro2022} serves as a prerequisite for defining strata, allowing us to categorize documents into coherent topics or themes. By sampling within each topic stratum, we ensure that the generated queries cover all relevant topics and variations within the knowledge base. 

Our five-step process for automatic query generation includes: (1) Embedding, (2) Dimensionality Reduction, (3) Clustering to Define Strata, (4) Stratified Sampling, and (5) LLM-driven Query Generation.

\bigskip
\ni{\bf Step 1: Embedding}\\
The first step involves generating embeddings for all documents in the collection. Embeddings are numerical vector representations that capture the semantic content of text; see \cite{Devin2019} for contextual embeddings using BERT (Bidirectional Encoder Representations from Transformers). In the proposed HCAT framework, we utilize the following advanced embedding models: 
\begin{itemize}
    \item {\sf Embeddings Trained through Contrastive Learning}: Models like SimCSE \citep{gao2021simcse} and Sentence-BERT \citep{RG2019} use contrastive learning to produce embeddings that capture fine-grained semantic similarities. These embeddings are effective for assessing the relevance and coherence between texts. See more discussion in Section~\ref{sec:funcmetrics}. 
    \item {\sf Specialized Embeddings from Natural Language Inference (NLI) Models}: NLI models are trained to determine entailment, contradiction, or neutrality between pairs of sentences \citep{Mac2009NLI}. By using embeddings from NLI models, we can evaluate the logical consistency and groundedness of the generated responses in relation to the source documents. Meanwhile, specialized NLI models can be applied for detecting hallucination \citep{KMXS2020, LHCB2022} and detecting toxicity \citep{JG2017, Hanu2020}. See more discussions in Sections~\ref{sec:funcmetrics} and \ref{sec:riskmetrics}.
\end{itemize}
By converting documents into embeddings, we create a foundation for analyzing semantic similarities or other discriminative tasks in a high-dimensional space.

\bigskip
\ni{\bf Step 2: Dimensionality Reduction}\\
The generated embeddings are high-dimensional vectors for which clustering approach may become less effective. To address this, we apply dimensionality reduction techniques to project the embeddings into a lower-dimensional space while preserving their essential semantic properties. Among others, we consider 
\begin{itemize}
    \item {\sf Principal Component Analysis (PCA)}: reduces dimensionality by linearly projecting data onto principal components that capture the most variance.
    \item {\sf Uniform Manifold Approximation and Projection (UMAP)}: preserves both local and global data structure, providing an efficient and scalable method for dimensionality reduction \citep{MHM2018}.
\end{itemize}
The choice of dimensionality reduction technique depends on factors such as dataset size, computational resources, and the desired balance between preserving local and global structures. It enables efficient clustering and visualization, facilitating the identification of natural groupings within the data.

\begin{figure}[htb!]
\centering
\includegraphics[width=0.8\textwidth]{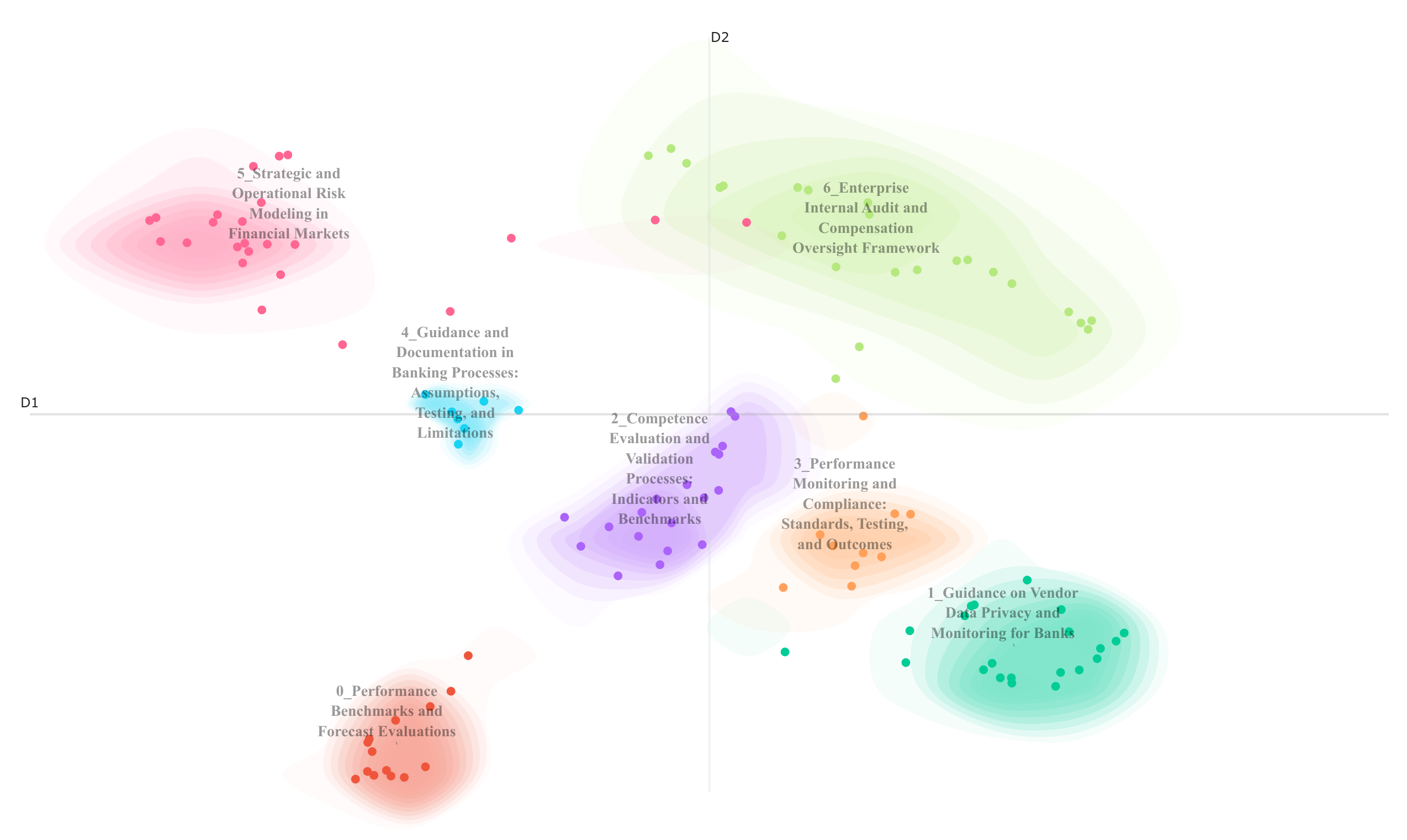}
\caption{Topic modeling through dimensionality reduction,  clustering, and topic extraction.}\label{fig:clustertopics}
\end{figure}

\bigskip
\ni{\bf Step 3: Clustering to Define Strata}\\
With the reduced-dimensional embeddings, we perform clustering to group semantically similar documents. A clustering algorithm may identify natural groupings within the data, effectively organizing the documents into topics or themes. Among others, we consider
\begin{itemize}
    \item {\sf K-Means Clustering}: Partitions the data into a predefined number of clusters by minimizing within-cluster variance.
    \item {\sf DBSCAN (Density-Based Spatial Clustering of Applications with Noise)}: Identifies clusters based on data point density, allowing for clusters of arbitrary shape and handling noise effectively \citep{schubert2017dbscan}.
\end{itemize}
The resulting clusters serve as strata for stratified sampling. Each cluster represents a distinct topic or sub-topic within the document collection, ensuring that all areas of the knowledge base are represented in the testing process. Figure~\ref{fig:clustertopics} shows an example of topic clustering, where each point in the plot represents a document chunk. This stratification is crucial for achieving comprehensive coverage and preventing biases toward dominant topics.

\bigskip
\ni{\bf Step 4: Stratified Sampling}\\
To achieve the comprehensive coverage effectively, we perform sampling within each cluster. Sampling can be proportional to the size of the cluster or weighted based on criteria such as the importance of the topic or the frequency of occurrence.

The stratified sampling approach ensures that queries are generated from all topics, preventing over-representation of prevalent themes and under-representation of niche areas. It allows for a balanced evaluation of the RAG system across the entire spectrum of the knowledge base, minimizing the risk of overlooking any significant areas.

\bigskip
\ni{\bf Step 5: LLM-driven Query Generation}\\
Finally, we utilize an LLM to generate queries based on the sampled documents. For each selected document, we prompt the LLM to create questions that are relevant to the content. The process involves:
\begin{enumerate}
    \item {\sf Extracting Key Information}: Identifying important facts, concepts, or statements within the document suitable for question formulation.
    \item {\sf Prompting the LLM}: Providing the LLM with the extracted information and instructions to generate queries of various types and complexities. 
    \item {\sf Ensuring Diversity and Complexity}: Instructing the LLM to produce a variety of question formats and difficulty levels, including yes/no questions, multiple-choice questions, and open-ended queries.
    \item {\sf Query Selection}: Evaluate and select queries based on relevancy metrics. 
\end{enumerate}

When prompting the LLM, we need to generate queries with various query types and complexities in order to thoroughly test the RAG system \citep{Yang2018Hot, RWGS2020, Li2024Auto}. Among our considerations are the following scenarios:
\begin{enumerate}
    \item {\sf Simple Factual Queries} that test basic retrieval capabilities.
    \item {\sf Multi-hop or Compound Queries} that assess the ability to synthesize information from multiple sources.
    \item {\sf Inference and Reasoning Queries} that evaluate logical and reasoning skills.
    \item {\sf Yes/No and Multiple-Choice Questions} that testing precision and understanding.
\end{enumerate}

To sum up, the automatic test generation component of the HCAT framework ensures that GLMs are tested comprehensively and representatively. The five-step process covers the entire scope of the document collection in a RAG system. By automating the query generation process using an LLM, we efficiently create a comprehensive set of test queries that are diverse in content and form. Through the use of topic modeling and clustering, it allows us to thoroughly evaluate the capabilities of a RAG system in retrieving relevant information and generating accurate responses across all topics.

\section{Functionality Metrics}\label{sec:funcmetrics}
Evaluating GLMs has traditionally involved metrics like BLEU, ROUGE, and perplexity, which quantify aspects of language generation such as n-gram overlap and fluency. However, these metrics often fail to capture semantic relevance and do not align well with human judgments, especially for open-ended generation tasks.
Recent research has explored embedding-based metrics that assess semantic similarity, offering a closer approximation to human judgments \citep{ZKWW2020}. 

We advocate the use of embedding-based evaluation metrics.
As discussed in Section~\ref{sec:autotest}, the embeddings of documents can be trained by contrastive learning or extracted from specialized NLI models. Using these embeddings, we can calculate semantic similarities and entailment probabilities, providing transparent and statistically grounded evaluation metrics. This approach avoids reliance on black-box tools or unverified methods, allowing for in-depth analysis and understanding of the evaluation results.

\begin{figure}[htb!]
\centering
\includegraphics[width=0.8\textwidth]{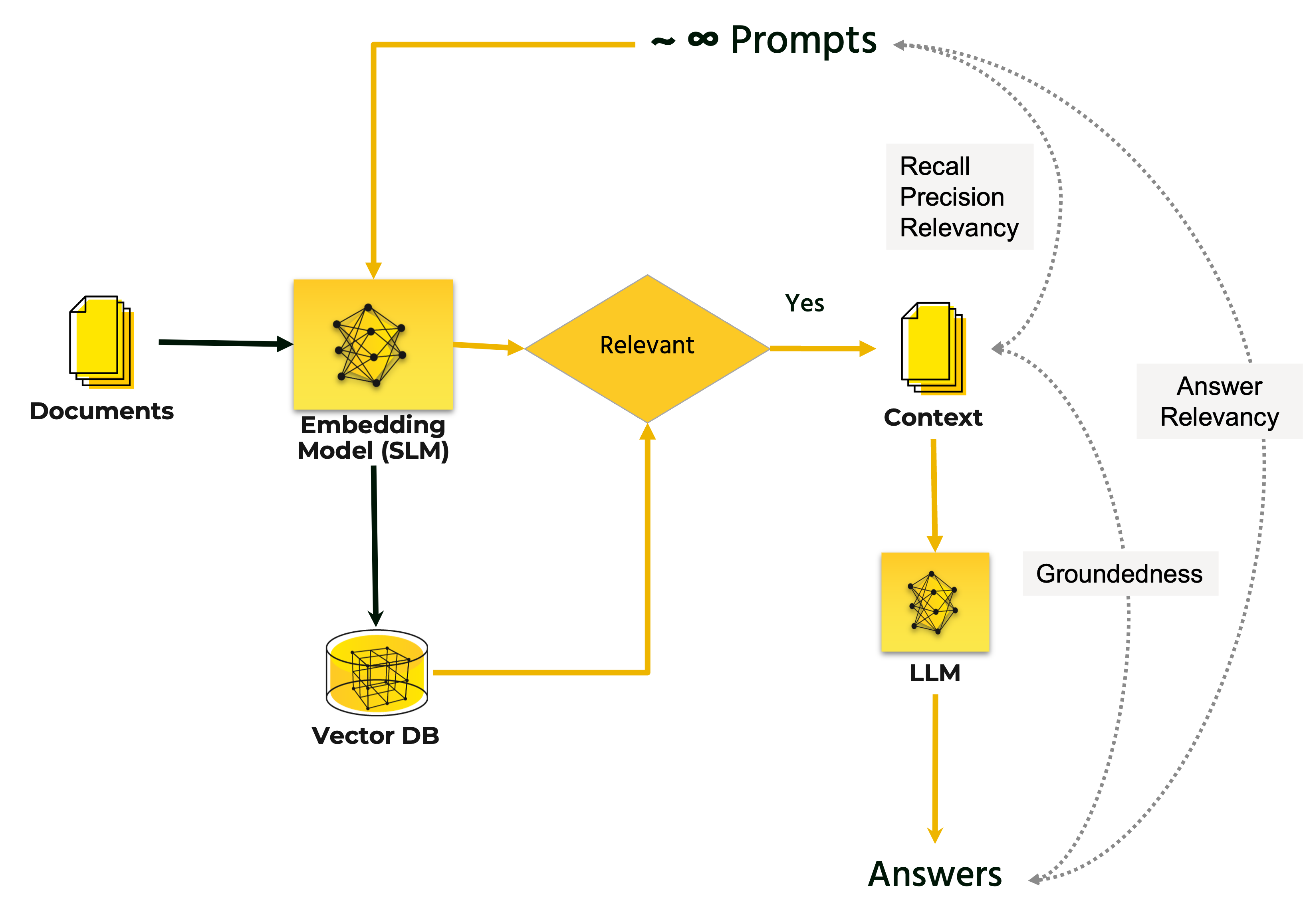}
\caption{RAG System Components and Functionality Evaluation}\label{fig:RAGeval}
\end{figure}

To effectively measure the performance of RAG systems, it is essential to adopt evaluation approaches that are both transparent and explainable, particularly when assessing retrieval relevance, groundedness, completeness, and answer relevancy; see Figure~\ref{fig:RAGeval} for an illustration.  The transparency is crucial not only for applications in regulated industries, where compliance and accountability are paramount, but also for fostering trustworthiness among users. Moreover, explainable metrics can be effectively calibrated with human evaluations, ensuring that automated assessments align with human judgments and expectations. By prioritizing explainable and interpretable evaluation methods, we can enhance the reliability and integrity of RAG systems, ensuring they meet high standards of performance and user trust.

\subsection{Context Relevancy}
Context relevancy measures how well the retrieved documents 
address the input query for a RAG system. To quantitatively measure the relevancy between a query and a context in RAG systems, we develop a sentence-level semantic similarity approach that extends the token-level approach from \cite{ZKWW2020}. This method breaks down both the query and the context into individual sentences and computes similarity scores for each pair, providing a fine-grained assessment of context relevancy. 

Let us denote the query as $\mathcal{Q} = \{ q_1, q_2, \dots, q_m \}$ consisting of $m$ sentences, and the retrieved context as $\mathcal{C} = \{ c_1, c_2, \dots, c_n \}$ consisting of $n$ sentences. For each sentence in either $\mathcal{Q}$ or $\mathcal{C}$, its embedding is computed using a suitable embedding model:
  \[
  \mathbf{e}_{q_i} = \text{Embed}(q_i), \quad \text{for } i = 1, 2, \dots, m.
  \]
  \[
  \mathbf{e}_{c_j} = \text{Embed}(c_j), \quad \text{for } j = 1, 2, \dots, n.
  \]
where $\text{Embed}(\cdot)$ represents the embedding function that maps a sentence to a vector in the $d$-dimensional embedding space. Thus, we can compute the similarity between each pair of query and context sentences using the {\em cosine similarity},
\[
\text{Sim}(q_i, c_j) = \cos(\theta_{ij}) = \frac{ \mathbf{e}_{q_i} \cdot \mathbf{e}_{c_j} }{ \| \mathbf{e}_{q_i} \| \| \mathbf{e}_{c_j} \| },
\]
where $\mathbf{e}_{q_i} \cdot \mathbf{e}_{c_j}$ is the dot product of two embedding vectors, and $\|\mathbf{e}_{q_i}\|, \|\mathbf{e}_{c_j}\|$ are the norms of the embeddings. The cosine similarity ranges from \(-1\) to \(1\), where \(1\) indicates identical orientation (maximum similarity), \(0\) indicates orthogonality (no similarity), and \(-1\) indicates opposite orientation.

Based on the cosine similarity for a pair of sentences, we may calculate the maximum similarity for each query sentence \( q_i \) by
\[
S_{\text{max}}(q_i) = \max_{1 \leq j \leq n} \text{Sim}(q_i, c_j).
\]
This \( S_{\text{max}}(q_i) \) score represents how well the query sentence \( q_i \) is addressed by the most relevant context sentence. Then, to measure the overall context relevancy of the whole query, we may aggregate the maximum similarity scores for all query sentences, i.e., 
  \[
  S_{\text{c-relevancy}} = \frac{1}{m} \sum_{i=1}^{m} S_{\text{max}}(q_i).
  \]
If certain query sentences are more important, we may use the weighted average,
  \[
  S_{\text{c-relevancy}} = \sum_{i=1}^{m} w_i \cdot S_{\text{max}}(q_i),
  \]
where $w_i$ is the weight assigned to the query sentence \( q_i \) subject to $w_i \geq 0$ and $\sum_{i=1}^{m} w_i = 1$. Furthermore, when it is critical that all aspects of the query should be addressed, we may use the minimax score that focuses on the least addressed query sentence, 
  \[
  S_{\text{c-relevancy}} = \min_{1 \leq i \leq m} S_{\text{max}}(q_i).
  \]
A high $S_{\text{c-relevancy}}$ score indicates that the context $\mathcal{C}$ is highly relevant to the query $\mathcal{Q}$, while a low 
$S_{\text{c-relevancy}}$ score indicates low relevancy. 
    
\subsection{Groundedness} 
Groundedness ensures that the generated content is based on the retrieved documents, avoiding unsupported statements or hallucinations. To measure the groundedness between the context and the generated answer in a RAG system, we employ two approaches: sentence similarity and natural language inference. 

\subsubsection{Sentence Similarity}
Denote the answer as $\mathcal{A} = \{a_1, a_2, \dots, a_k\}$ consisting of $k$ sentences.
Similar to the approach of computing the context relevancy, let us break down both the context and the answer into individual sentences, then compute the similarity scores for each pair of sentence embeddings. For each sentence $a_i$ in the answer, calculate the maximum similarity by
$$
S_{\text{max}}(a_i)  = \max_{1 \leq j \leq n} \text{Sim}(a_i, c_j),
$$
which measures how well the answer sentence \(a_i\) is grounded in the context $\mathcal{C} = \{ c_1, c_2, \dots, c_n \}$. To obtain an overall groundedness score for the entire answer, we may aggregate the maximum similarity scores for all answer sentences: 
\[   
S_{\text{groundedness}}= \frac{1}{k} \sum_{i=1}^{k} S_{\text{max}}(a_i). 
\]
A high $S_{\text{groundedness}}$ score indicates that on average the sentences in the answer are well-supported by the context, suggesting that the answer is grounded and less likely to contain hallucinations.
 
Conversely, a low $S_{\text{groundedness}}$ score suggests that some sentences in the answer may lack sufficient support from the context. The sentence with the lowest similarity to any context sentence, identified as
\[ 
i^* = \arg \min_{1 \leq i \leq k} S_{\text{max}}(a_i), 
\]
is considered the least grounded and may indicate a potential hallucination. Other sentences with low $S_{\text{max}}(a_i)$ values could also signal possible hallucinations.

\subsubsection{Natural Language Inference (NLI)}
NLI models are specifically designed to determine the inferential relationship between two pieces of text, a premise and a hypothesis, by classifying the relationship as ``entailment'', ``neutral'', or ``contradiction'' \citep{Mac2009NLI}. In the context of RAG systems, we treat the context as the premise and the generated answer as the hypothesis.

NLI aims to determine whether a hypothesis can logically be inferred from a premise. For the purpose of measuring groundedness (the opposite of hallucination), NLI provides a mechanism to assess whether the generated answer is logically supported by the context. If the answer is entailed by the context, it is considered grounded; if it contradicts the context or is unrelated, it may indicate a hallucination.

While NLI models provide class probabilities through multi-class classification, we can obtain a more nuanced groundedness measure by analyzing the embeddings produced by the model and measuring the distance to the decision boundary. The decision boundary in the embedding space separates different classes and reflects the model's confidence in its predictions. In this method, the distance to the decision boundary is directly related to the logit value for the ``entailment'' class. This approach simplifies the calculation and provides an interpretable measure of groundedness. 

Suppose a linear classifier computes a logit score \( z \) using the embedding \( \mathbf{x} \) of the input (which combines the premise and hypothesis):
\[
z = \mathbf{w}^\top \mathbf{x} + b
\]
where \( \mathbf{w} \) is the weight vector and  \( b \) is the bias term.  The distance \( D \) from the input point \( \mathbf{x} \) to the decision boundary (i.e., the hyperplane \( \mathbf{w}^\top \mathbf{x} + b = 0 \)) is given by:
\[
D = \frac{\mathbf{w}^\top \mathbf{x} + b}{\|\mathbf{w}\|} = \frac{z}{\|\mathbf{w}\|}
\]
where \( \|\mathbf{w}\| \) is the Euclidean norm (magnitude) of the weight vector. The distance to the decision boundary can be used to measure the groundedness in the sense that 
\begin{itemize}
        \item when $D>0$, the hypothesis is on the entailment side, i.e., grounded; 
        \item when $D<0$, the hypothesis is on the non-entailment side, i.e., potential hallucination.
\end{itemize}
We can map the distance \( D \) to a probability groundedness score between 0 and 1 by applying the logit transformation $\sigma(D) = 1/(1 + e^{-D})$.

In sentence-level groundedness assessment, each sentence $a_i$ in the answer $\mathcal{A}$ is combined with the context $\mathcal{C}$, then input into the NLI model to obtain the embedding \( \mathbf{x}_i \). This allows us to compute the sentence-level $z_i,D_i$ and $\sigma(D_i)$. The sentences with low $\sigma(D_i)$ scores may be identified as potential hallucinations.

\subsection{Completeness} 
Completeness evaluates whether the generated answer covers all relevant information from the context. In RAG systems, completeness refers to the degree to which the generated answer incorporates all relevant information from the retrieved context. A complete answer should not only be accurate and relevant but also cover all essential points that are pertinent to the user's query. Ensuring completeness is crucial for providing users with comprehensive and informative responses.

\subsubsection{Sentence Similarity}
This approach assesses completeness by evaluating how well the sentences in the context are reflected in the generated answer. Similar to context relevancy and groundedness, we break down the context and answer into sentences, then calculate embedding-based similarity in the sentence level. For each sentence \( c_i \) in the context $\mathcal{C} = \{ c_1, c_2, \dots, c_n \}$, calculate the maximum similarity with any sentence in the answer $\mathcal{A} = \{a_1, a_2, \dots, a_k\}$:
   \[
   S_{\text{max}}(c_i) = \max_{1 \leq j \leq k} \text{Sim}(c_i, a_j).
   \]
This score indicates how well the context sentence \( c_i \) is covered in the answer.

Similarly, we can aggregate $S_{\text{max}}(c_i)$ scores to obtain the overall completeness score by either the simple average
\[
S_{\text{completeness}} = \frac{1}{n} \sum_{i=1}^{n} S_{\text{max}}(c_i),
\]
or the weighted average
\[
S_{\text{completeness}} = \sum_{i=1}^{n} w_i \cdot S_{\text{max}}(c_i), 
\]
where $w_i$ is the weight assigned to the context sentence \( c_i \) subject to $w_i \geq 0$ and $\sum_{i=1}^{n} w_i = 1$. A high completeness score indicates that the answer covers most of the content from the context, while a low completeness score suggests that significant portions of the context are not reflected in the answer.

\subsubsection{Distribution Alignment Using Wasserstein Distance}
When assessing the completeness of an answer generated by LLMs, it is essential to measure how well the answer captures the entire information distribution of the original context. The sentence similarity approach focuses on finding close matches between individual sentences, which may not fully reflect the answer’s coverage of the overall context. By applying {\em Wasserstein distance}, a measure from optimal transport theory, we can evaluate the alignment of information distribution between the context and the summary, offering a complementary perspective on completeness; see also \cite{TangOpt2022}.
 
Optimal transport \citep{ChewiOpt2024} is a mathematical approach for measuring the cost of transforming one distribution into another. Wasserstein distance, also known as Earth Mover’s distance, is an optimal transport metric that quantifies the minimum cost to align two distributions, reflecting how closely they match in structure and content. In the context of evaluating RAG-generated answers, 
the context sentences are treated as a distribution of information that needs to be represented in the answer sentences. In this case, 
Wasserstein distance measures how much effort is required to transform the distribution of context information $\mathcal{C}$ into the distribution of answer information $\mathcal{A}$:
\[
W(\mathcal{C}, \mathcal{A}) = \min_{\gamma \in \Gamma(p, q)} \sum_{i=1}^{n} \sum_{j=1}^{k} \gamma_{ij} d(c_i, a_j)
\]
where \( d(c_i, s_j) \) is the distance (e.g., Euclidean or cosine distance) between the embeddings of context sentence \( c_i \) and answer sentence \( a_j \), \( \gamma_{ij} \) is the transport weight that represents how much of context sentence \( c_i \) is mapped to answer sentence \( a_j \), and \( \Gamma(p, q) \) is the set of all possible transport plans. 

The goal is to find the transport plan \( \gamma \) that minimizes the total cost, yielding the optimal Wasserstein distance between the distributions.
For simplicity, assume each sentence contributes equally to the overall information distribution by assigning each sentence a weight of \( 1/n \) for context sentences and \( 1/k \) for answer sentences. Then, calculate the average transport cost, weighted by \( 1/nk \) based on the uniform weights. This involves summing the pairwise distances and dividing by the total number of pairs, 
\[
W(\mathcal{C}, \mathcal{A}) = \frac{1}{nk} \sum_{i=1}^{n} \sum_{j=1}^{k} d(c_i, a_j).
\]
This average distance provides a straightforward approximation of the Wasserstein distance.

To interpret the Wasserstein distance for completeness,  a lower distance indicates that the summary effectively captures the distribution of information in the context, suggesting higher completeness. A higher Wasserstein  distance implies gaps in completeness, where the answer may be missing critical content from the context.

\subsubsection{Complementing Sentence Similarity with Wasserstein Distance}
While sentence similarity methods directly compare individual pairs of sentences, Wasserstein distance provides a complementary approach by assessing the global alignment of information across the entire context and answer. The key advantages of Wasserstein distance are:
\begin{enumerate}
\item {\sf Global Information Distribution}: Wasserstein distance captures the overall distribution of information in the context and answer, making it effective for assessing completeness by reflecting how well the summary represents the main ideas and topics from the context.
\item {\sf Tolerance to Partial Matches}:
Unlike sentence similarity, which requires exact or near-exact matches, Wasserstein distance allows for approximate alignment. This means that summaries with paraphrased or generalized content can still achieve low Wasserstein distances, provided they retain the main themes.
\item {\sf Contextual Relationships}: 
By aligning entire sentence distributions, Wasserstein distance indirectly accounts for the thematic structure of the context and summary, offering a broader perspective than sentence-level similarity alone.
\item {\sf Evaluating Completeness}:
A lower Wasserstein distance indicates that the answer closely captures the distribution of information from the context, reflecting higher completeness. Conversely, a higher distance suggests missing content or inadequate coverage of essential topics.
\end{enumerate}

Using both sentence similarity and Wasserstein distance together allows for a more nuanced evaluation of completeness, capturing both the detailed alignment of specific sentences and the overall distribution of ideas and topics within the answer.

\subsection{Answer Relevancy} 
Answer relevancy ensures that the generated response directly addresses the user's query. We calculate maximum similarity scores between the query and the generated response to measure alignment with the user's intent. Again, we can employ the sentence-level similarity approach similar to previously discussed metrics. This time we break down both the query and the answer into individual sentences and computing similarity scores for each pair. By analyzing these similarities, we can quantify how well the answer addresses the user's query.

For each sentence \(a_i\) in the answer $\mathcal{A} = \{a_1, a_2, \dots, a_k\}$,  calculate the maximum similarity with any sentence in the query $\mathcal{Q} = \{ q_1, q_2, \dots, q_m \}$:
\[ 
S_{\text{max}}(a_i) = \max_{1 \leq j \leq m} \text{Sim}(a_i, q_j), 
\]
which measures how well the answer sentence \(a_i\) is relevant to the query. Then, to obtain an overall answer relevancy score for the entire answer, we can aggregate the $S_{\text{max}}(a_i)$ scores for all answer sentences by either the simple average
\[ 
S_{\text{a-relevancy}} = \frac{1}{k} \sum_{i=1}^{k} S_{\text{max}}(a_i)
\]
or the weighted average
\[ 
S_{\text{a-relevancy}} = \sum_{i=1}^{k} w_i \cdot S_{\text{max}}(a_i), 
\] 
where \(w_i\) is the weight assigned to the answer sentence \(a_i\) subject to $w_i \geq 0$ and $\sum_{i=1}^{k} w_i = 1$.
A high $S_{\text{a-relevancy}}$ score indicates that on average the answer effectively addresses the user's question, while a low $S_{\text{a-relevancy}}$ score indicates potential divergence or irrelevance.

When it is critical that all parts of the answer are desired to be relevant to the query, we may focus on the least relevant answer sentence
and compute the minimum of maximum similarities:
\[ 
S_{\text{a-relevancy}} = \min_{1 \leq i \leq k} S_{\text{max}}(a_i). 
\]

\section{Risk and Safety Metrics}\label{sec:riskmetrics}
In this section, we provide a concise overview of evaluation metrics for assessing risk and safety aspects of generative language models. Here we focus on critical dimensions that ensure the deployed GLMs perform responsibly, and in alignment with regulatory standards for appropriate use in high-stakes applications such as banking. Three essential risk and safety metrics for validating GLMs in these environments include toxicity assessment, bias evaluation, and privacy protection. 

{\bf Toxicity assessment} measures the likelihood that a model generates harmful, offensive, or inappropriate content.  In banking, where interactions must be professional and respectful, toxicity in model outputs can severely damage customer trust, harm the institution's reputation, and potentially lead to legal repercussions if sensitive or controversial topics are mishandled. Toxicity is typically assessed through NLI models that classify statements as safe or offensive, allowing for real-time toxicity assessment; see  \cite{Hanu2020} among others. 

{\bf Bias evaluation} focuses on detecting demographic or sentiment bias within the model responses, ensuring equitable treatment of diverse user groups. In banking, where interactions may influence financial decisions or customer perceptions, bias can lead to discriminatory responses, harming the reputation of the institution and potentially leading to regulatory scrutiny. Bias evaluation involves testing the model with a diverse set of demographic-related queries, including variations in race, gender, age, and income level, to determine if response quality or sentiment varies across different groups. Sentiment analysis models help identify potential differences in tone or attitude, while counterfactual evaluation assesses whether altering demographic-related terms (e.g., swapping ``man'' with ``woman'') results in consistent responses. Models are scored on bias based on thresholds that align with banking standards, allowing institutions to identify and mitigate unacceptable biases before deployment. If biases are identified, they can be addressed by fine-tuning the model with additional data that represents underrepresented groups fairly, or by implementing constraints to reduce unintended biases in generated content.

{\bf Privacy protection} is essential to ensure the model does not disclose sensitive information, such as personal financial details, customer identities, or other proprietary data. In banking, privacy is paramount due to stringent regulatory requirements like the GDPR and CCPA, and privacy violations can result in severe financial and legal consequences. Privacy protection for GLMs often involves Named Entity Recognition (NER), which identifies and flags sensitive entities in the output, such as names, addresses, or account numbers, allowing the model to suppress or filter such information before it reaches the user. Additionally, contextual data filtering is implemented to identify phrases related to account transactions or other sensitive areas, reducing the risk of unintentional data leakage. Adversarial testing is used to simulate scenarios where users might try to elicit sensitive information, and the model responses are evaluated to ensure that they avoid privacy violations. Strict thresholds are set to flag any instance of sensitive information being revealed, prompting immediate action to investigate and adjust the model if necessary. Privacy safeguards are reinforced through model retraining or the implementation of guardrails, ensuring compliance with industry regulations and protecting customer information.

Together, these metrics form a robust framework for risk and safety validation of GLMs, particularly in banking. Each metric provides a unique lens for assessing the model behavior, guiding institutions in minimizing risks associated with inappropriate content, unfair treatment, and data privacy. This validation process helps ensure that models meet the high standards of accuracy, transparency, and accountability required in financial services, supporting safe and responsible deployment in real-world applications.

\section{Calibration of Machine and Human Evaluations}\label{sec:calib}
\begin{figure}[htbp!]
\centering
\smallskip
\begin{tikzpicture}
    \node[draw, fill=gray!20, minimum width=3cm, minimum height=1.5cm] at (0,4) (machine) {Machine-Generated Scores};
    \node[draw, fill=gray!20, minimum width=3cm, minimum height=1.5cm] at (0,0) (human) {Human Judgments};
    \draw[->, line width=0.4mm, shorten >=2pt, shorten <=2pt] (machine) -- (human);
    \node[right] at (0.5, 2.3) {Stage 1: Probability Calibration};
    \node[right] at (0.5, 1.7) {Stage 2: Conformal Prediction};
\end{tikzpicture}
\caption{Calibration Diagram of Machine and Human Evaluations}\label{fig:calib}
\end{figure}
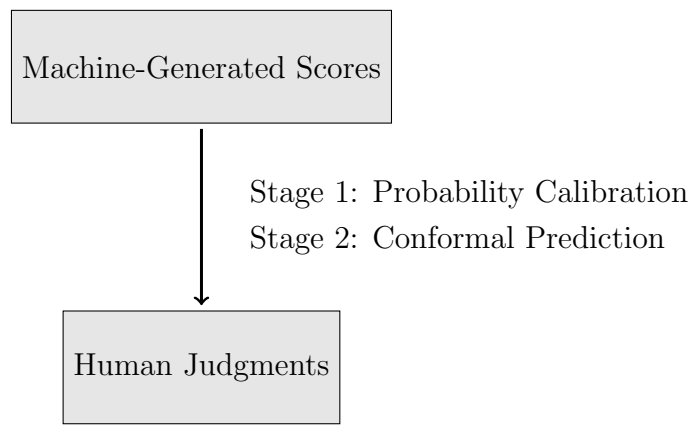

To ensure alignment between machine-generated scores and human judgments, we employ a double-calibration approach. This process consists of two stages: {\em probability calibration} and {\em conformal prediction}, each playing a unique role in producing reliable machine evaluations. Since conformal prediction adds an additional layer of calibration, we refer to this two-stage process as the double-calibration method. Figure~\ref{fig:calib} provides a diagram of this two-stage approach, with further details described below:
\begin{itemize}
\item Stage 1: Probability calibration provides an initial mapping of machine scores to probabilities that align with human expectations, 
\item Stage 2: Conformal prediction quantifies the uncertainty of these calibrated probabilities, providing prediction intervals with confidence levels. 
\end{itemize}

This double-calibration strategy allows us to link machine evaluation metrics (such as relevancy, groundedness, or completeness) to human evaluations that may be binary or multi-level. Error analysis (Type I and II errors) can be applied to assess the alignment of machine versus human evaluations. By setting calibrated thresholds on calibration model outputs, we derive prediction sets that reflect the evaluation confidence, facilitating both automated decision-making and human-in-the-loop processes.

\subsection{Stage 1: Probability Calibration}
The first stage of probability calibration aims to map raw machine-generated scores to calibrated probabilities that align with human judgments. This step translates machine evaluations into a probability scale, making it easier to interpret machine-generated scores in terms of human expectations.

To achieve this, we employ standard probability calibration methods, which include:
\begin{itemize}
\item {\sf Logistic Model}: For human evaluations with binary labels, a logistic regression model is used to map machine scores to probabilities, producing a sigmoid-shaped probability curve. This approach is also known as Platt scaling, and it is effective when machine scores have a roughly linear relationship with human judgments. For multi-category evaluations, it could involve probabilities assigned to each category.

\item {\sf Ordinal Regression}: For human evaluations with rankings, ordinal regression provides a calibrated probability for each level, reflecting the likelihood of each judgment category.

\item {\sf Isotonic Regression or Monotonic XGBoost}: These non-parametric methods provide a flexible, piecewise-constant mapping of scores to probabilities, especially useful when there is a monotonic relationship between machine scores and human evaluations.
\end{itemize}
Using a hold-out calibration dataset with both machine scores and human labels, we train the chosen calibration model to learn the mapping from raw scores to probabilities. Once trained, the model can be applied to any new machine-generated score, providing a calibrated probability that represents the likelihood of human agreement. 

\subsection{Stage 2: Conformal Prediction}
Conformal prediction is a flexible statistical framework that produces prediction intervals or sets for any model, assuming only data exchangeability \citep{VGS2005, AB2021}. It enables the creation of confidence sets that include the true label with a specified confidence level, making it well-suited to classification tasks with complex or uncertain mappings between machine evaluation metrics and human labels. In this paper, we consider only the split conformal prediction, while leaving the full conformal prediction to a future paper. 

In our context, conformal prediction applies a second level of calibration to the calibrated probabilities obtained from Stage~1. To implement this, we use an independent hold-out calibration sample, distinct from the sample used for probability calibration. This separate dataset includes machine-generated scores and corresponding human labels, allowing us to quantify the uncertainty around the calibrated probabilities established in Stage~1.

We discuss the split conformal prediction for the logistic calibration model $\hat{f}(\cdot)$. In the procedure below, \( S(x, y, \hat{f}) \) denotes the non-conformity score, which measures the deviation between the calibrated probability and the human evaluation. This notation is distinct from other uses of \( S \) in this paper.
\begin{enumerate}
    \item {\sf Non-Conformity Score Calculation}: For each sample in the independent calibration set with sample size $n$, compute the non-conformity score 
    \[ 
    S(x, y, \hat{f}) = |y - \hat{f}(x)| = 1 - P(Y=y|x),
    \]
    where \( \hat{f}(x) \)  is the calibrated probability from Stage 1, and \( y \) is the observed human label 1 or 0. 
    
    \item {\sf Calibrated Quantile Computation}: Choose a desired confidence level, represented by an error rate \( \alpha \) (e.g., \( \alpha = 0.1 \) for 90\% confidence). Then compute the quantile \( \hat{q} \) of the non-conformity scores as:
    \[
    \hat{q} = \text{Quantile} \left( \{S_1, S_2, \dots, S_n\}; \frac{\lceil(n + 1)(1 - \alpha)\rceil}{n+1} \right).
    \]

    \item {\sf Prediction Set Construction}: For a new test sample \( x_{\text{test}} \), create the prediction set \( \mathcal{T}(x_{\text{test}}) \) by including all possible values \( y \) such that:
    \[
    \mathcal{T}(x_{\text{test}}) = \{ y : S(x_{\text{test}}, y, \hat{f}(x_{\text{test}})) \leq \hat{q} \} = \big\{y: P(Y=y|x_{\text{test}}) \geq 1 - \hat{q}\big\}.
    \]
\end{enumerate}

\begin{figure}[hbt!]
\centering
\includegraphics[width=0.8\textwidth]{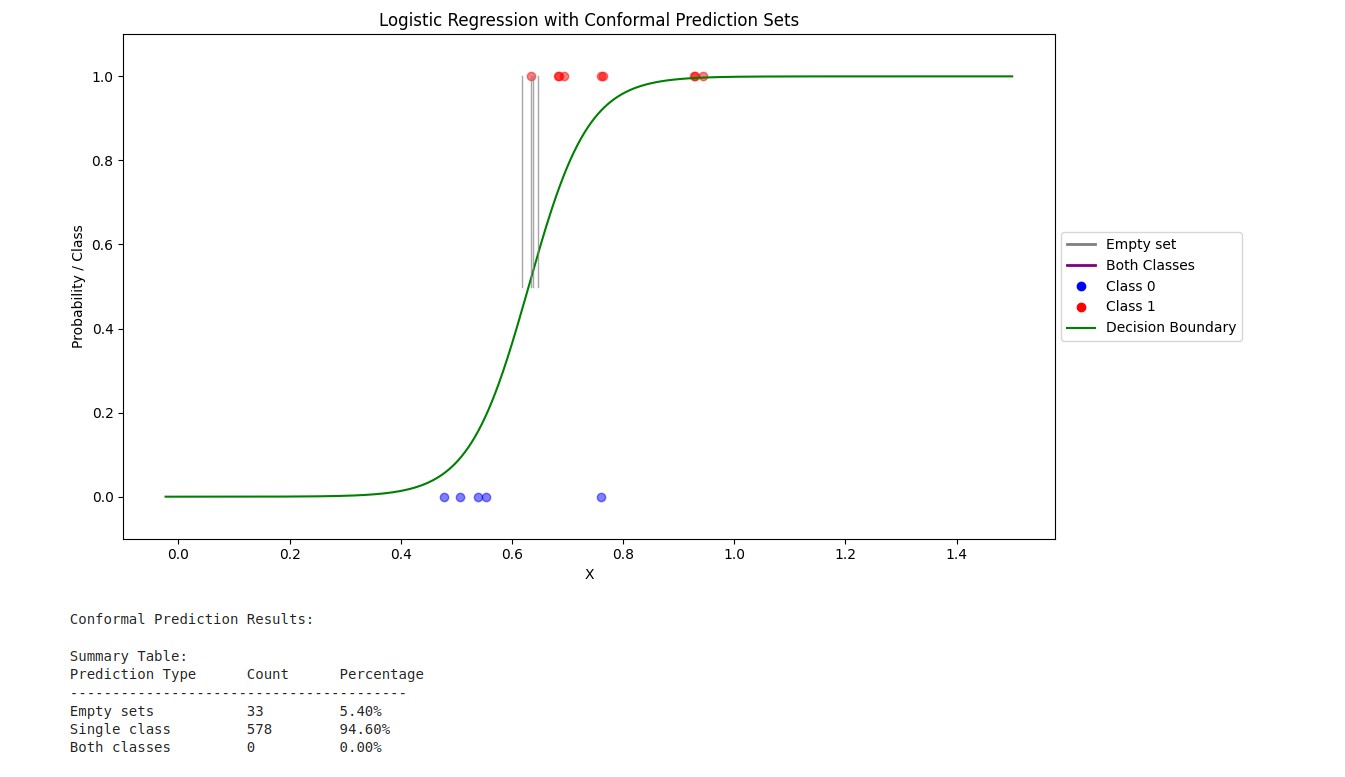}
\caption{An illustration of calibration for machine-human groundedness evaluation,  using logistic regression and conformal prediction.}\label{fig:LogitConf}
\end{figure}

For each new instance $x_{\rm test}$, the prediction set is among the three possible cases depending on the predicted probability $\hat{f}(x_{\text{test}})$ and the calibrated quantile $\hat{q}$. 
\begin{itemize}
    \item {\sf Single-Class Set}: \{0\} when $\hat{f}(x_{\text{test}}) < \min\{\hat{q}, 1-\hat{q}\}$ ,  or \{1\} when $\hat{f}(x_{\text{test}}) > \max\{\hat{q}, 1-\hat{q}\}$.
    \item {\sf Both-Class Set}: \{0, 1\} when $1-\hat{q}\leq \hat{f}(x_{\text{test}}) \leq \hat{q}$ and $\hat{q}\geq 0.5$.
    \item {\sf Empty Set}: $\emptyset$ when $\hat{q} < \hat{f}(x_{\text{test}}) <1-\hat{q}$ and $\hat{q} < 0.5$.
\end{itemize}
The single-class prediction set indicates high confidence that the true label corresponds to this class. For instance, a prediction set at 90\% confidence level suggests a high likelihood that the true evaluation aligns with this class. On the other hand, when the prediction set includes both classes or empty, it reflects uncertainty and suggests that additional review may be needed. Figure~\ref{fig:LogitConf} presents an example of machine-human calibration using logistic regression for groundedness evaluation with binary labels from human judgment.

\section{Robustness and Weakness Analysis}\label{sec:robustweak}
This section provides an overview of two essential perspectives in the testing and validation of GLMs: robustness testing and weakness identification. Robustness testing examines the stability and resilience of the model under varied input conditions, while weakness identification aims to pinpoint specific areas where the model may need improvement. Together, these evaluations offer insights into the model’s strengths and potential vulnerabilities, guiding enhancements before deployment in critical applications.
 
\subsection{Robustness Testing}
Robustness testing assesses the ability of a RAG system to handle diverse and challenging inputs (including potentially problematic inputs), ensuring that it performs robustly across a range of real-world scenarios. The testing includes three main types of robustness checks:
\begin{enumerate}
    \item {\sf Adversarial Inputs}: This involves introducing deliberately misleading or contradictory information to the model’s input. Adversarial testing can expose how the model handles conflicting information and whether it can distinguish relevant content from distractors. By subjecting the model to adversarial queries, evaluators can assess its ability to avoid generating inaccurate or biased responses, which is particularly important in regulated industries like banking where factual accuracy is critical.

    \item {\sf Out-of-Distribution Queries}: To examine the model’s adaptability, robustness testing includes queries on topics that are not present within the model’s training data or document collection. Out-of-distribution queries help reveal the model’s limitations by testing how it responds to unfamiliar topics. A robust RAG model should either respond appropriately based on the closest relevant information or acknowledge its limitations instead of generating inaccurate information.

    \item {\sf Input Variations}: The system is also evaluated on its ability to handle input variations, including spelling errors, grammatical mistakes, and colloquial language. This type of testing is essential for real-world applications where user inputs may be unstructured or contain errors. A resilient RAG system should be able to interpret such variations and generate coherent responses despite minor input inaccuracies.
\end{enumerate}

By systematically examining the model  performance across these query types, the robustness testing phase highlights potential weaknesses, guiding targeted improvements. This ensures that the model remains reliable and resilient across various practical scenarios and input complexities, enhancing its robustness for deployment in critical applications.

\subsection{Weakness Identification}
\begin{figure}[htbp!]
\centering
\includegraphics[width=0.87\textwidth]{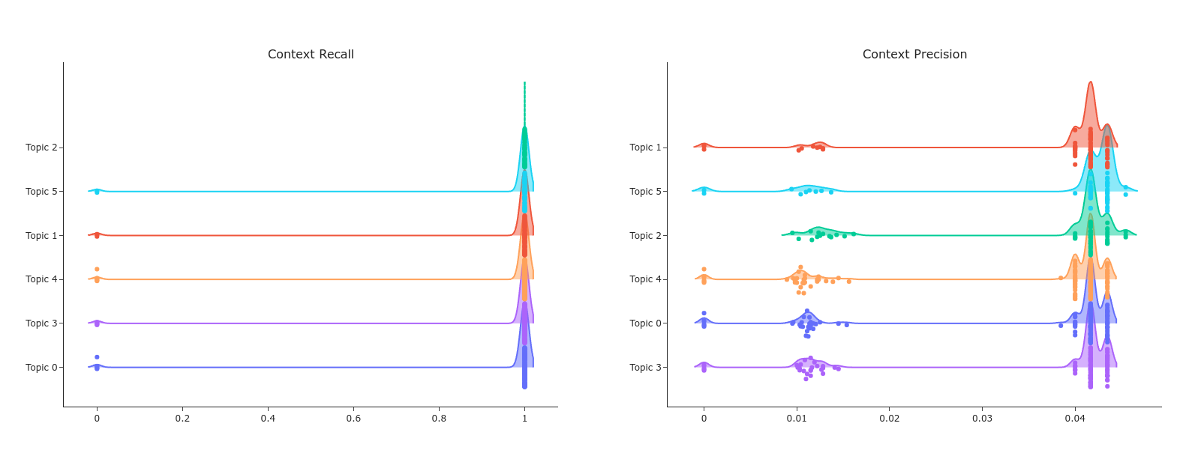}
\caption{Marginal (topic) weakness analysis: recall \& precision}\label{fig:Weak1}
\end{figure}

\begin{figure}[htbp!]
\centering
\bigskip
\includegraphics[width=0.87\textwidth]{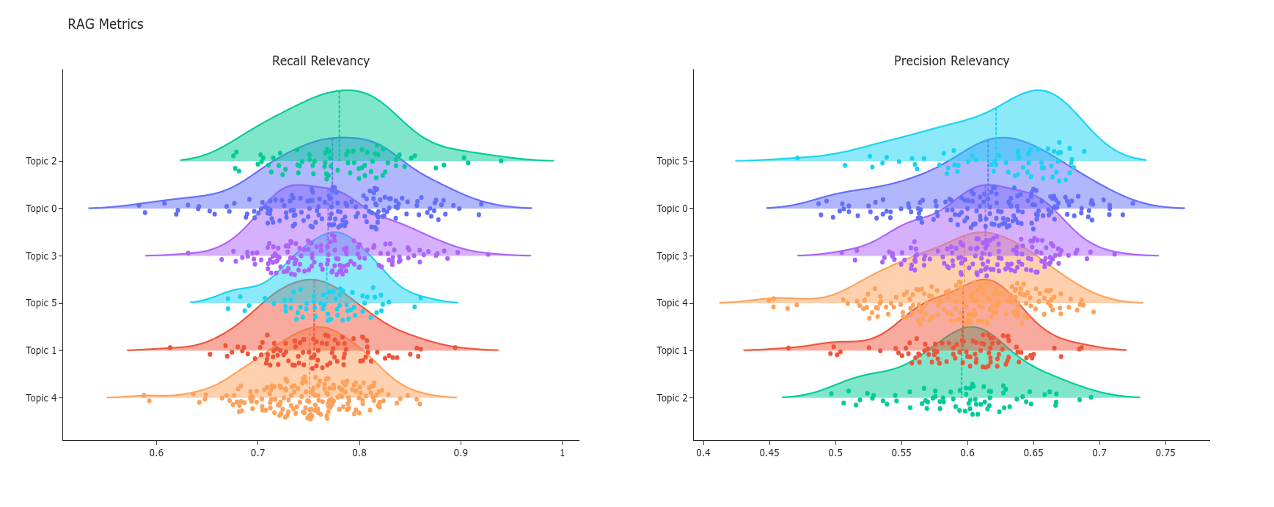}
\caption{Marginal (topic) weakness analysis: relevancy (sentence recall \& precision)}\label{fig:Weak2}
\end{figure}

\begin{figure}[htbp!]
\centering
\bigskip
\includegraphics[width=0.87\textwidth]{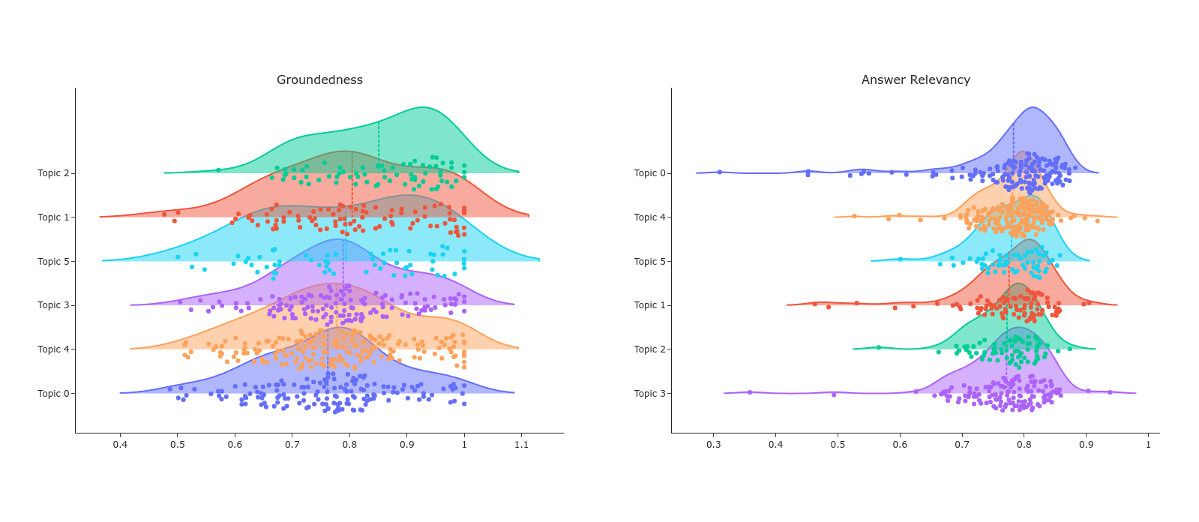}
\caption{Marginal (topic) weakness analysis: groundedness \& answer relevancy}\label{fig:Weak3}
\end{figure}

Model weakness identification, provides an approach for pinpointing specific performance issues within a RAG system. This process allows for a granular understanding of the model's areas of under performance, supporting targeted improvements. The methodology in weakness identification includes the following techniques:
\begin{enumerate}
    \item {\sf Marginal Analysis}: It is a key technique where the system’s performance is evaluated across individual dimensions, such as topics or query types.  By breaking down metrics for each category, marginal analysis helps identify specific areas where the model does not perform well. For instance, by analyzing relevance or groundedness scores for each topic individually, the evaluation can reveal topics that require further data augmentation or model fine-tuning. Marginal analysis is critical for finding isolated weaknesses that might be masked in aggregate evaluations. Figures~\ref{fig:Weak1}--\ref{fig:Weak3} show examples of marginal plots of various metrics. Each point is the evaluation result from each query, low values are where the weakness are.
    
    \item {\sf Bivariate Analysis}: This method examines the interaction between two dimensions, such as topic and query type, to uncover joint weaknesses. This approach is beneficial for identifying compound issues that may not be apparent when looking at a single dimension alone. For example, a model may perform adequately on simple questions within a topic but struggle with more complex, multi-hop questions in the same area. This joint examination helps in pinpointing specific combinations that are challenging for the model and might benefit from additional training or refinement strategies.

    \item {\sf Visualization Techniques}: Tools such as heatmap and violin plots represent performance distributions across different metrics, making it easier to communicate and interpret weaknesses. For example, heatmaps can visually indicate low-performance areas in terms of recall, precision, or relevance, allowing stakeholders to identify specific problem areas quickly. Visualization supports the practical implementation of weakness identification by clearly showing areas where performance dips, guiding model improvement initiatives in an accessible format.
\end{enumerate}

Together, these methods provide a strategy for detailed, data-driven analysis of model weaknesses, guiding focused enhancement efforts to optimize the RAG system's effectiveness and reliability before deployment. 

\section{Discussion and Conclusion}\label{sec:con}
This paper has presented a HCAT framework for testing and validation of GLMs, specifically tailored for RAG systems used in high-stakes applications like banking. By leveraging the structured, bounded nature of RAG systems, where generation is constrained by a defined document collection, we address the inherent challenges of evaluating GLMs in open-ended domains. Our proposed framework provides a robust, scalable, and transparent solution for testing GLMs, integrating multiple validation methodologies to enhance model reliability, interpretability, and compliance.

The framework presented here offers several key benefits:
\begin{itemize}
    \item {\sf Comprehensive Testing}: Automatic query generation through stratified sampling based on topic modeling ensures thorough coverage of the knowledge base. This systematic approach allows us to generate diverse and representative queries that test the model across all relevant topics and query types.

    \item {\sf Explainable Evaluation Metrics}: By employing embedding-based evaluation metrics, including embeddings trained through contrastive learning and specialized embeddings from NLI models, we provide transparent and interpretable assessments. This avoids reliance on black-box tools and allows for in-depth analysis of semantic relevance, groundedness, and logical consistency between queries, documents, and responses.

    \item {\sf Trustworthy Evaluations}: Calibration with human judgments aligns machine evaluations with human perceptions, enhancing the reliability of automated assessments. By accounting for the limitations of algorithmic metrics, we ensure that the evaluation reflects qualities valued by users. 

    \item {\sf Robustness Assessment}: It evaluates the model stability against varied scenarios, such as adversarial inputs, out-of-distribution queries, and linguistic variations, ensuring reliable performance in diverse practical contexts. This comprehensive robustness testing uncovers vulnerabilities, ensuring consistent performance under challenging conditions.

    \item {\sf Targeted Improvements}: Weakness identification through marginal and bivariate analysis enables focused enhancements. By pinpointing specific topics or query types where the model underperforms, we can prioritize areas for improvement and optimize the system's overall effectiveness.
\end{itemize}

While the framework addresses many challenges, caution must be taken due to certain limitations such as:
\begin{itemize}
    \item {\sf Quality of Topic Modeling}: The effectiveness of stratified sampling depends on the accuracy of topic modeling. Inaccurate or overly broad topics may lead to insufficient coverage or imbalanced sampling. 
    \item {\sf Human Calibration Sample Size}: The representativeness of human evaluations may be limited by sample size and diversity. Expanding the calibration dataset with more extensive and diverse human judgments can improve alignment between machine evaluations and human perceptions, leading to more accurate calibration.
    \item {\sf Evolving Language Models}: As generative language models continue to advance, evaluation methods must adapt accordingly. Ongoing research into embedding techniques and evaluation metrics is necessary to keep pace with model developments and ensure the framework remains effective and relevant.
\end{itemize}

In conclusion, the proposed framework addresses the critical need for comprehensive and reliable evaluation of generative language models, particularly within the context of RAG systems. By systematically combining automatic test generation, explainable and human-calibrated evaluation metrics, robustness testing, and targeted weakness identification, we enhance the trustworthiness of these models.


\begin{thebibliography}{99}
\bibitem[Angelopoulos and Bates, 2023]{AB2021}
Angelopoulos, A. N. and Bates, S. (2023). Conformal prediction: A gentle introduction. {\em Foundations and Trends in Machine Learning}, {\bf 16}(4), pp. 494--591.

\bibitem[Brown et~al., 2020]{brown2020language}
Brown, T., Mann, B., Ryder, N., Subbiah, M., Kaplan, J. D., Dhariwal, P., \ldots, and Amodei, D. (2020). Language models are few-shot learners. \textit{Advances in Neural Information Processing Systems}, {\bf 33}, pp. 1877--1901.

\bibitem[Chewi et~al., 2024]{ChewiOpt2024}
Chewi, S.,  Niles-Weed, J. and Rigollet, P. (2024).  Statistical optimal transport. \textit{arXiv preprint: 2407.18163}

\bibitem[Devlin et~al., 2019]{Devin2019}
Devlin, J., Chang, M. W., Lee, K. and Toutanova, K. (2019). BERT: Pre-training of deep bidirectional transformers for language understanding. 
{\em Proceedings of the 2019 Conference of the North {A}merican Chapter of the Association for Computational Linguistics: Human Language Technologies, Volume 1 (Long and Short Papers)}, pp. 4171--4186.

\bibitem[Gao et~al., 2021]{gao2021simcse}
Gao, T., Yao, X. and Chen, D. (2021), 
SimCSE: Simple contrastive learning of sentence embeddings.
\textit{Proceedings of the 2021 Conference on Empirical Methods in Natural Language Processing},  pp. 6894--6910.

\bibitem[Grootendorst, 2022]{Gro2022}
Grootendorst, M. (2022). BERTopic: Neural topic modeling with a class-based TF-IDF procedure. \textit{arXiv preprint: 2203.05794}.

\bibitem[Hanu and Unitary team, 2020]{Hanu2020}
Hanu, L. and Unitary team. (2020). Detoxify. \textit{https://github.com/unitaryai/detoxify}

\bibitem[Jigsaw and Google, 2017]{JG2017}
Jigsaw and Google. (2017). Perspective API. \textit{https://www.perspectiveapi.com/}

\bibitem[Khashabi et~al., 2021]{khashabi2021qafacteval}
Khashabi, D., Stanovsky, G., Bragg, J., Lourie, N., Kasai, J., Choi, Y.,\ldots, and Weld, D. (2021).
QAFactEval: Improved QA-based factual consistency evaluation.
\textit{arXiv preprint: 2112.08542}.

\bibitem[Kryscinski et~al., 2020]{KMXS2020}
Kryscinski, W., McCann, B., Xiong, C. and Socher, R. (2020). Evaluating the factual consistency of abstractive text summarization. \textit{arXiv preprint: 2004.04228}.


\bibitem[Krishna et~al., 2023]{krishna2023rag}
Krishna, K., Khattab, G., Moschella, R., Zhao, P. and Zhang, C. (2023).
RAG: A comprehensive survey of retrieval-augmented text generation.
\textit{arXiv preprint: 2312.10997}.

\bibitem[Laban et~al., 2022]{LHCB2022}
Laban, P., Hirst, L., Cummings, J. and Bansal, M. (2022). SummaC: Re-visiting NLI-based models for inconsistency detection in summarization. \textit{Transactions of the Association for Computational Linguistics}, 10, pp. 163--177.

\bibitem[Lewis et~al., 2020]{lewis2020retrieval}
Lewis, P., Perez, E., Piktus, A., Petroni, F., Karpukhin, V., Goyal, N., \ldots, and Kiela, D. (2020).
Retrieval-augmented generation for knowledge-intensive nlp tasks.
\textit{Advances in Neural Information Processing Systems}, {\bf 33}, pp. 9459--9474.

\bibitem[Liang et~al., 2023]{Liang2023HELM}
Liang, P., Bommasani, R., Lee, T., Tsipras, D., Soylu, D., Yasunaga, M., \ldots, and Koreeda, Y. (2023). Holistic evaluation of language models. \textit{Transactions on Machine Learning Research}.


\bibitem[Lin et~al., 2022]{Lin2022TruthQA}
Lin, S., Hilton, J. and Evans, O. (2022). TruthfulQA: Measuring how models mimic human falsehoods. {\em Proceedings of the 60th Annual Meeting of the Association for Computational Linguistics},  Volume 1: Long Papers (pp. 3214--3252).

\bibitem[Li et~al., 2024]{Li2024Auto}
Li, Y., Singh, R., Joshi, T. and  Sudjianto,A. (2024). Automatic generation of behavioral test cases for natural language processing using clustering and prompting. \textit{arXiv preprint: 2408.00161}.

\bibitem[MacCartney, 2009]{Mac2009NLI}
MacCartney, B. (2009). {\em Natural Language Inference}. Doctoral Dissertation, Stanford University.

\bibitem[McInnes et~al., 2018]{MHM2018}
McInnes, L., Healy, J. and Melville, J. (2018). UMAP: Uniform manifold approximation and projection for dimension reduction. \textit{arXiv preprint: 1802.03426}.

\bibitem[Reimers and Gurevych, 2019]{RG2019}
Reimers, N. and Gurevych, I. (2019). Sentence-BERT: Sentence embeddings using Siamese BERT-networks. \textit{Proceedings of the 2019 Conference on Empirical Methods in Natural Language Processing and the 9th International Joint Conference on Natural Language Processing (EMNLP-IJCNLP)}, pp. 3982–-3992.

\bibitem[Ribeiro et~al., 2020]{RWGS2020}
Ribeiro, M. T., Wu, T., Guestrin, C. and Singh, S. (2020). Beyond accuracy: behavioral Testing of NLP models with checkList. \textit{Proceedings of the 58th Annual Meeting of the Association for Computational Linguistics}, pp. 4902--4912.

\bibitem[Schubert et~al., 2017]{schubert2017dbscan}
Schubert, E., Sander, J., Ester, M., Kriegel, H. P., and Xu, X. (2017). DBSCAN revisited, revisited: Why and how you should (still) use DBSCAN. {\em ACM Transactions on Database Systems}, 42(3), pp. 1--21.

\bibitem[Sudjianto and Zhang, 2024]{sudjianto2024MVP}
Sudjianto, A. and Zhang, A. (2024). Model validation practice in banking: A structured approach for predictive models. Available at \textit{SSRN: https://ssrn.com/abstract=4977043}.

\bibitem[Srivastava et~al., 2023]{Srivastava2023}
Srivastava, A., Rastogi, A., Rao, A., Shoeb, A. A., Abid, A., Fisch, A., \ldots, and Mehta, H. (2023). Beyond the imitation game: Quantifying and extrapolating the capabilities of language models. \textit{Transactions on Machine Learning Research}.

\bibitem[Tang et~al., 2022]{TangOpt2022}
Tang, P., Hu, K., Yan, R., Zhang, L., Gao, J. and Wang, Z. (2022). OTExtSum: Extractive text summarisation with optimal transport. {\em  Findings of the Association for Computational Linguistics: NAACL 2022},  pp. 1128--1141.


\bibitem[Vovk et~al., 2005]{VGS2005}
Vovk, V., Gammerman, A. and Shafer, G. (2005). Algorithmic learning in a random world. \textit{Springer Science \& Business Media}.

\bibitem[Yang et~al., 2018]{Yang2018Hot}
Yang, Z. Qi, P., Zhang, S., Bengio, Y., Cohen, W.W., Salakhutdinov, R. and Manning, C.D. (2018). HotpotQA: A dataset for diverse, explainable multi-hop question answering. \textit{arXiv Preprint: 1809.09600}.

\bibitem[Zhang et~al., 2020]{ZKWW2020}
Zhang, T., Kishore, V., Wu, F., Weinberger, K. Q. and Artzi, Y. (2020). BERTScore: Evaluating text generation with BERT. \textit{International Conference on Learning Representations}.

\bibitem[Zhao et~al., 2023]{Zhao202survey}
Zhao, W. X., Zhou, K., Li, J., Tang, T., Wang, X., Hou, Y., \ldots, and Wen, J. R. (2023). A survey of large language models. {\em arXiv preprint: 2303.18223}.

\end{thebibliography}
\end{document}